\documentclass[runningheads]{llncs}
\usepackage[T1]{fontenc}
\usepackage{booktabs}
\usepackage{multirow}
\usepackage{amsmath}
\usepackage{amssymb}
\usepackage{hyperref}
\usepackage{graphicx,verbatim}
\begin{document}

\titlerunning{CellVTA: Enhancing Vision Foundation Models for Cell Segmentation}
\title{CellVTA: Enhancing Vision Foundation Models for Accurate Cell Segmentation and Classification
}


%

\author{Yang Yang$^1$, Xijie Xu$^1$, Yixun Zhou$^2$, Jie Zheng$^{1,3,*}$}  
\authorrunning{Yang et al.}
\institute{$^{1}$ ShanghaiTech University, $^{2}$ Shanghai Ocean University,\\
           $^{3}$ Shanghai Engineering Research Center of Intelligent Vision and Imaging \\
    \email{zhengjie@shanghaitech.edu.cn}}

\maketitle              
\begin{abstract}
Cell instance segmentation is a fundamental task in digital pathology with broad clinical applications.  Recently, vision foundation models, which are predominantly based on Vision Transformers (ViTs), have achieved remarkable success in pathology image analysis. However, their improvements in cell instance segmentation remain limited. A key challenge arises from the tokenization process in ViTs, which substantially reduces the spatial resolution of input images, leading to suboptimal segmentation quality, especially for small and densely packed cells. To address this problem, we propose CellVTA (Cell Vision Transformer with Adapter), a novel method that improves the performance of vision foundation models for cell instance segmentation by incorporating a CNN-based adapter module. This adapter extracts high-resolution spatial information from input images and injects it into the ViT through a cross-attention mechanism. Our method preserves the core architecture of ViT, ensuring seamless integration with pretrained foundation models. Extensive experiments show that CellVTA achieves 0.538 mPQ on the CoNIC dataset and 0.506 mPQ on the PanNuke dataset, which significantly outperforms the state-of-the-art cell segmentation methods. Ablation studies confirm the superiority of our approach over other fine-tuning strategies, including decoder-only fine-tuning and full fine-tuning. Our code and models are publicly available at \url{https://github.com/JieZheng-ShanghaiTech/CellVTA}.

\keywords{Cell instance segmentation, foundation model, computational pathology}

\end{abstract}
\section{Introduction}

Cell instance segmentation is a fundamental task in digital pathology, which is critical for cancer diagnosis and treatment~\cite{binder2021morphological,wang2022cell}. It involves the precise delineation of cell boundaries and classification of cell types. Many deep learning methods have been proposed to tackle this problem. Convolutional neural networks (CNNs) are the most commonly used methods in this task, such as Hover-Net~\cite{graham2019hover} and Micro-Net~\cite{raza2019micro}. This kind of method demonstrates strong performance, as their architectures capture local spatial structures,  which is an effective inductive prior for image-based tasks. Recently, foundation models have achieved remarkable success in natural language processing~\cite{achiam2023gpt} and have become increasingly popular in computer vision~\cite{kirillov2023segment}, known as vision foundation models (VFMs). VFMs have shown excellent performance in most areas of computational pathology~\cite{chen2024towards}, such as tumor classification and tissue segmentation. Their success can be attributed to greater model capacity of the Transformer architecture~\cite{vaswani2017attention} behind VFMs, which allows them to gain rich prior knowledge through large-scale pretraining on extensive pathology datasets.

\begin{figure}[t]
\centering
\includegraphics[width=0.9\textwidth]{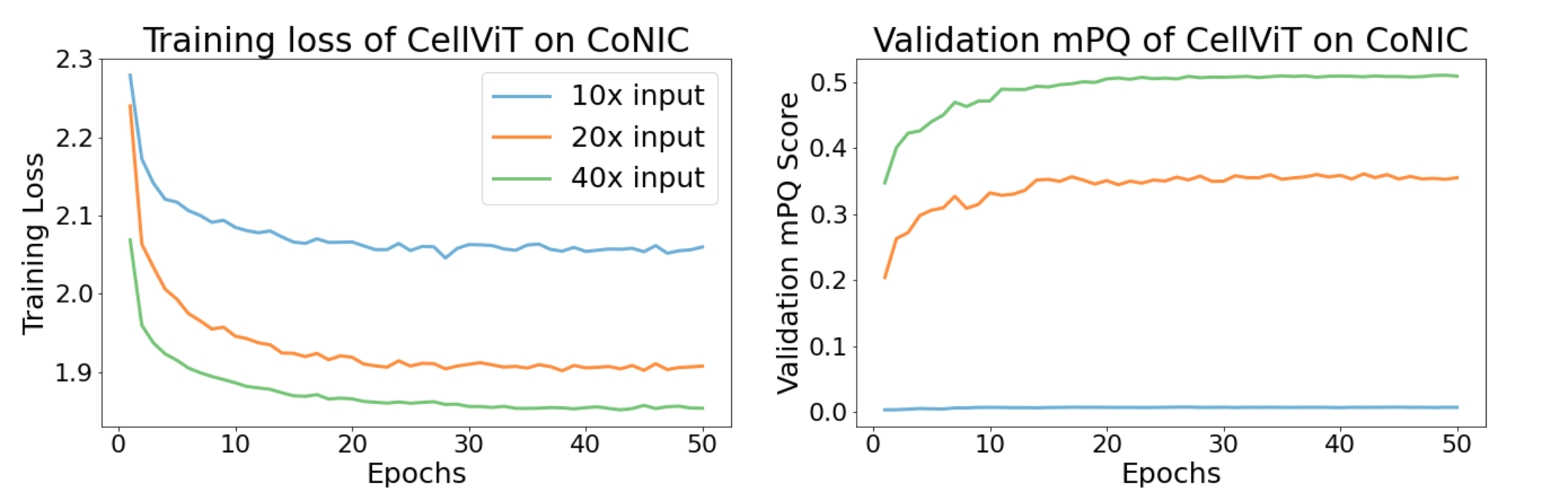}
\caption{Training loss and validation mPQ of CellViT under different magnifications. We use 2$\times$ downsampling and upsampling to generate 10$\times$ and 40$\times$ input images.} \label{cellvit_conic_loss}
\end{figure}


While VFMs achieve high performance in many computational pathology tasks, their improvements in cell segmentation remain limited~\cite{chen2024towards,stringer2024transformers,vadori2025mind}. A key challenge stems from the architecture of Vision Transformers (ViTs),  which serve as the backbone of most VFMs. In pathology images, cells are often small and densely packed. Standard ViTs employ a patch-based tokenization process that typically downsamples the input image by a factor of 16, yielding patch sizes comparable to individual cells. Such aggressive reduction in spatial resolution will significantly degrade segmentation quality.  As shown in Fig.~\ref{cellvit_conic_loss}, the cell segmentation quality of CellViT~\cite{horst2024cellvit} on the CoNIC dataset~\cite{graham2024conic} significantly drops when the magnification becomes smaller. Another limitation is that standard ViTs lack image-specific inductive biases, which results in slower convergence and lower performance, compared to CNNs~\cite{raghu2021vision}. These two challenges are mainly caused by the structures of standard ViTs. A potential solution is to modify their architecture. Indeed, many variants of ViT have been proposed and have achieved better segmentation performance~\cite{liu2021swin,wang2021pyramid}. However, modifying the structure of ViT would hinder the utilization of VFMs, as most VFMs are based on standard ViTs. Therefore, our goal is to enhance the performance of VFMs in cell segmentation while preserving the standard ViT architecture.

To address this problem, we propose CellVTA (Cell Vision Transformer with Adapter), a novel method that improves the performance of VFMs for cell instance segmentation in pathology images by incorporating multi-scale spatial features through a CNN-based adapter. This adapter integrates local and fine-grained details into the ViT’s feature representations via a cross-attention mechanism, without modifying the core ViT architecture. This injection of multi-scale information significantly augments the model’s sensitivity to small and densely packed objects, thereby improving its performance in the cell segmentation task. We conduct extensive experiments on the CoNIC~\cite{graham2024conic} and PanNuke~\cite{gamper2020pannuke} datasets, which are the most challenging cell segmentation datasets across multiple organs and cell types. The results demonstrate that our model achieves 0.538 mPQ on CoNIC and 0.506 mPQ on PanNuke, which significantly outperforms the state-of-the-art (SOTA) methods. Ablation studies show that the strategy of CellVTA achieves better performance than decoder-only fine-tuning and full fine-tuning.

\begin{figure}[t]
\includegraphics[width=\textwidth]{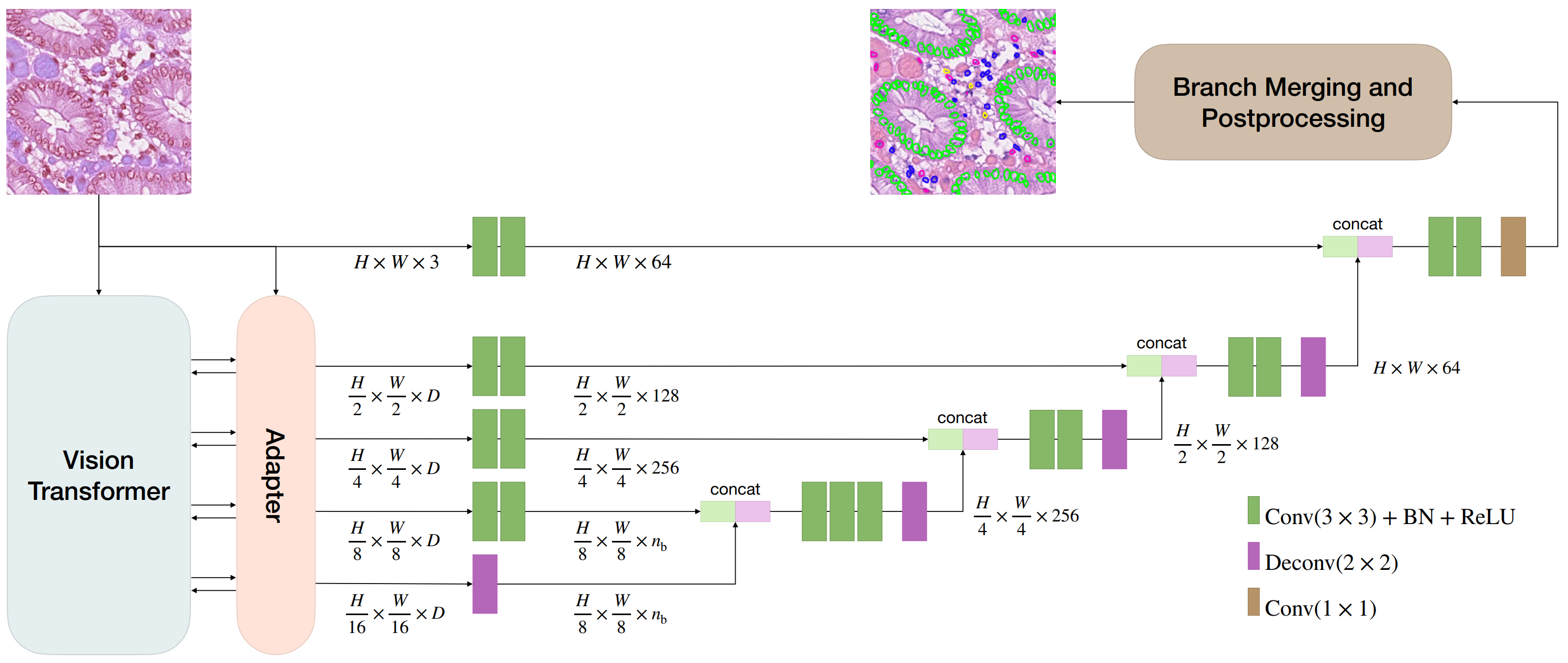}
\caption{Overall architecture of CellVTA. It comprises: (1) a ViT encoder, (2) an adapter module, and (3) a multi-branch decoder. First, the ViT encoder extracts features from an input image. Then the adapter module extracts multi-scale spatial information from the input image and injects them into the ViT encoder via feature interaction. The outputs of adapter are passed to the decoder via skip connections for cell segmentation.} \label{cellvit_adapter_structure}
\end{figure}

\section{Method}
\subsubsection{Overall Architecture.} As shown in Fig.~\ref{cellvit_adapter_structure}, CellVTA consists of three main components: (1) a ViT encoder, (2) an adapter module, and (3) a multi-branch decoder. Our model builds upon the CellViT~\cite{horst2024cellvit} framework, which employs a standard ViT encoder, making it well-suited for leveraging vision foundation models (VFMs), such as SAM~\cite{kirillov2023segment} and UNI~\cite{chen2024towards}. Inspired by ViT-adapter~\cite{chen2023vision}, we design an adapter module to extract high-resolution spatial information from input images via CNNs and then inject it into the features of the ViT encoder via a cross-attention mechanism, which helps to restore fine-grained details lost during tokenization. This enhancement is the key innovation of our approach. 


\begin{figure}[t]
\includegraphics[width=\textwidth]{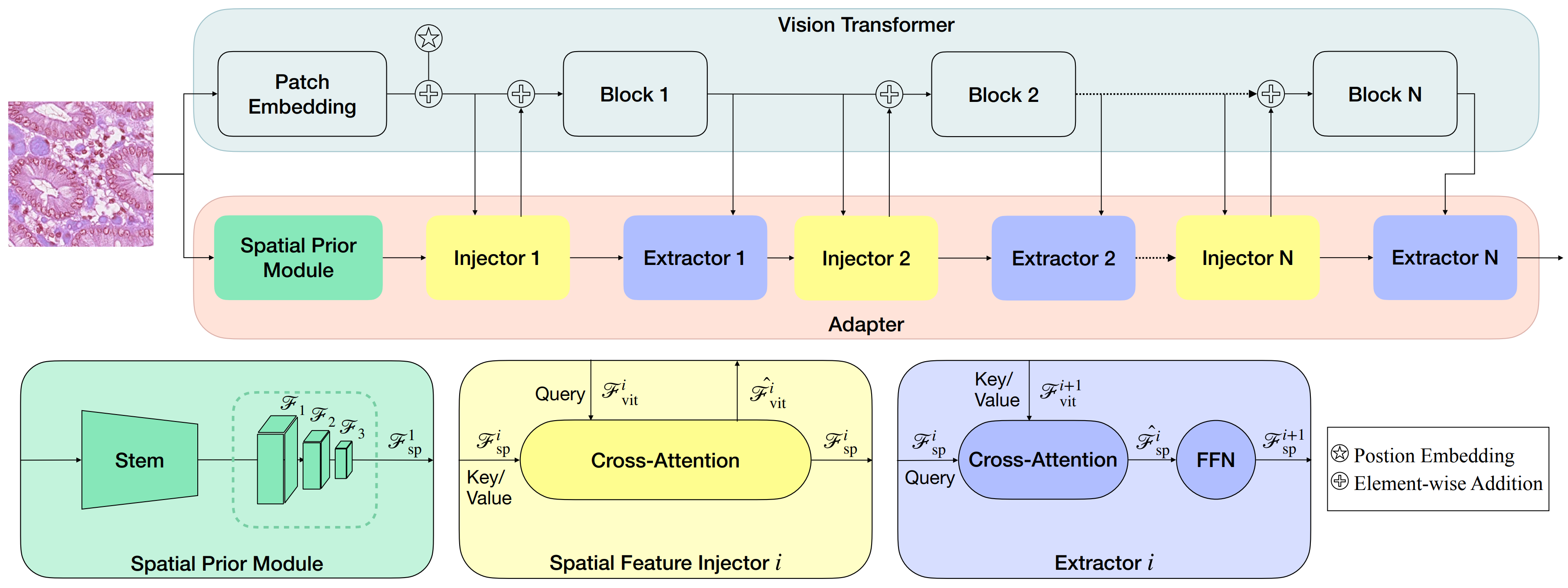}
\caption{Detailed architecture of the adapter module. The upper branch is the ViT encoder which is divided into $N$ ($N = 4$ in this paper) equal blocks for feature interaction. The lower branch is the adapter module consisting of  (1) a spatial prior module to extract high-resolution spatial features from input images (2) a spatial feature injector to inject spatial priors into the ViT (3) a multi-scale feature extractor to extract hierarchical information from the ViT features. } \label{adapter_detail}
\end{figure}

\subsubsection{ViT Encoder.}  In the ViT encoder, the input images $x\in \mathbb{R}^{H\times W \times C}$ are divided into a sequence of flattened tokens $x_p \in \mathbb{R}^{N\times P^2\cdot C}$, where ($H$, $W$) is the image resolution and $C$ is the number of channels. Each token is an image patch with dimension ($P$, $P$) and $N=HW/P^2$ is the number of resulting tokens. Then the flattened tokens are projected to a $D$-dimensional space with a trainable linear layer $E$. Additionally, a learnable 1D position embedding $E_\mathrm{pos}$~\cite{dosovitskiy2021an}  and a class token $x_\mathrm{class}$ are added to form the final input of the Transformer encoder, which can be formulated as: $z_0 = [x_\mathrm{class};x_\mathrm{p}^{1}E;x_\mathrm{p}^{2}E;...;x_\mathrm{p}^{N}E] + E_\mathrm{pos}$. The Transformer encoder consists of $L$ Transformer blocks with multihead self-attention ($\mathrm{MHA}$) and multi-layer perceptron ($\mathrm{MLP}$) layers. Layer normalization ($\mathrm{LN}$) and residual connections are used. A latent vector $z_i$ in each block is calculated by:
\begin{equation}
    z_{i}' = \mathrm{MHA}(\mathrm{LN}(z_{i-1})) + z_{i-1}, i=1\dots L
\end{equation}
\begin{equation}
    z_{i} = \mathrm{MLP}(\mathrm{LN}(z_{i}')) + z_{i}', i=1\dots L
\end{equation}


\noindent\textbf{Adapter Module.} Inspired by~\cite{chen2023vision},  we design an Adapter Module integrated with the ViT encoder. Fig.~\ref{adapter_detail} shows the detailed structure of the adapter. It comprises three components: (1) a spatial prior module (SPM) to extract high-resolution spatial features from input images; (2) a spatial feature injector to inject spatial priors into the ViT encoder; (3) a multi-scale feature extractor to extract hierarchical information from the ViT features. 

We use a CNN to serve as the SPM, which consists of four convolutional blocks. The first block contains four convolutional layers, while the others have two convolutional layers, with the last layer of each block at stride 2 and the rest at stride 1. A $1 \times 1$ convolution is used to map the feature maps to $D$ dimension. Thus, we get a feature pyramid $\{\mathcal{F}_1, \mathcal{F}_2, \mathcal{F}_3\}$ from the last three blocks, which contains feature maps with a resolution of 1/4, 1/8, and 1/16 of the input image. Then they are flattened and concatenated into a sequence of tokens $\mathcal{F}_\mathrm{sp}^{1}\in \mathbb{R}^{(\frac{HW}{8^2}+\frac{HW}{16^2}+\frac{HW}{32^2})\times D}$, as input to the spatial feature injector. 

The spatial feature injector uses cross-attention to inject spatial priors  $\mathcal{F}_\mathrm{vit}^{i}$ into the ViT feature $\mathcal{F}_\mathrm{vit}^{i}$ at the $i$-th block, with $\mathcal{F}_\mathrm{vit}^{i}$ as query and $\mathcal{F}_\mathrm{sp}^{i}$ as key and value:
\begin{equation}
    \hat{\mathcal{F}}_{\mathrm{vit}}^{i} = \mathcal{F}_{\mathrm{vit}}^{i} + \gamma^{i} \mathrm{Attention}(\mathrm{LN}(\mathcal{F}_{\mathrm{vit}}^{i}), \mathrm{LN}(\mathcal{F}_{\mathrm{sp}}^{i}))
\end{equation}
where Deformable Attention~\cite{zhu2021deformable} is used as Attention($\cdot$) and a learnable vector $\gamma^i$ (initialized with 0) balances the output of the attention layer and ViT features.

 After injection, $\hat{\mathcal{F}}_{\mathrm{vit}}^{i}$ are fed into the $i$-th block and the output is $\mathcal{F}_{\mathrm{vit}}^{i+1}$. Then we use a multi-scale feature extractor module, consisting of a cross-attention layer and a feed-forward network (FFN), to extract multi-scale features:
\begin{equation}
    \mathcal{F}_{\mathrm{sp}}^{i+1} = \hat{\mathcal{F}}_{\mathrm{sp}}^{i} + \mathrm{FFN}(\mathrm{norm}(\hat{\mathcal{F}}_{\mathrm{sp}}^{i}))
\end{equation}
\begin{equation}
    \hat{\mathcal{F}}_{\mathrm{sp}}^{i} = \mathcal{F}_{\mathrm{sp}}^{i} + \mathrm{Attention}(\mathrm{norm}(\mathcal{F}_{\mathrm{sp}}^{i}), \mathrm{norm}(\mathcal{F}_{\mathrm{vit}}^{i+1}))
\end{equation}
Here, the spatial feature $\mathcal{F}_{\mathrm{sp}}^{i}$ is query and the ViT feature $\mathcal{F}_{\mathrm{vit}}^{i+1}$ is key and value. The output $\mathcal{F}_{\mathrm{sp}}^{i+1}$ is used as the input of the next spatial feature injector. Finally, we build the 1/2-scale feature map by upsampling the 1/4-scale feature map via a deconvolutional layer. In this way, we get a feature pyramid $\{ h_1, h_2, h_3, h_4\}$ from the adapter module for decoding.

\subsubsection{Decoder and Skip Connections.} The decoder of CellVTA comprises three branches following HoverNet~\cite{graham2019hover}: the nuclear pixel (NP) branch for nuclei binary segmentation, the nuclear classification (NC) branch for nuclei type semantic segmentation, and the HoVer (HV) branch for predicting the horizontal and vertical distances of nuclear pixels to their centers of mass. In addition, our model adopts a U-Net structure that the encoder is connected to the decoders via five skip connections to leverage information at multiple encoder depths. The first skip connection processes the input image $x$ with a convolutional layer followed by batch normalization and ReLU. For the remaining four skip connections, the latent embeddings of the adapter module $h_i$ ($i=1,2,3,4$) are extracted and reshaped to 2D feature maps $H_i \in R^{\frac{H}{2^i}\times \frac{W}{2^i}\times D}$. Then each feature map is processed by convolutional layers for dimension adjustment (except for $H_4$ are 2$\times$ upsampled by a deconvolutional layer), and concatenated with the corresponding decoder features. Here, the shape of the encoder features exactly matches the corresponding decoder features. The class token $z_{L,\mathrm{class}}$ is used for tissue classification as an auxiliary task using a linear classifier.

\subsubsection{Optimization and Postprocessing.} We use the same loss function as CellViT~\cite{horst2024cellvit}: $\mathcal{L}_{\mathrm{total}} = \mathcal{L}_{\mathrm{NP}}  + \mathcal{L}_{\mathrm{HV}} + \mathcal{L}_{\mathrm{NC}} + \mathcal{L}_{\mathrm{TC}}$, where $\mathcal{L}_{\mathrm{NP}}$ consists of Dice loss and Focal Tversky (FT) loss~\cite{abraham2019novel}, $\mathcal{L}_{\mathrm{HV}}$ consists of MSE and MSGE loss, $\mathcal{L}_{\mathrm{NC}}$ consists of Dice, FT and cross entropy (CE) loss, and $\mathcal{L}_{\mathrm{TC}}$ loss is a CE loss. For inference, postprocessing follows~\cite{graham2019hover} to merge the outputs of three decoder branches to generate the final instance predictions with the watershed algorithm.

\section{Experiment}

\subsection{Experimental Setup}
\textbf{Datasets.} We perform comprehensive evaluations of CellVTA on two datasets: PanNuke~\cite{gamper2020pannuke} and CoNIC~\cite{graham2024conic}, which are two of the largest manually annotated cell segmentation datasets. PanNuke consists of 7,904 images ($256\times 256$ px)  across 19 tissue types, with 189,744 annotated nuclei from 5 cell types. The images are captured at a magnification of $40\times$ ($0.25$ $\mathrm{\mu m/px}$). CoNIC contains 4,891 colon images ($256\times 256$ px) with 495,179 annotated nuclei from 6 cell types, captured at $20\times$ magnification ($\sim0.5$ $\mathrm{\mu m/px}$). Both datasets are highly challenging due to their multi-tissue and multi-source composition, and severe class imbalance. For PanNuke, we follow the three-fold cross-validation splits provided by the PanNuke dataset organizers~\cite{gamper2020pannuke} and report the averaged results over three splits. For CoNIC, we split it into training set and test set by patients with a ratio of $8:2$, and further split 20\% of the training set as validation set.

\begin{table}[t]
\centering
\setlength{\tabcolsep}{6pt}
\caption{Performance comparison between CellVTA and baselines on CoNIC. Top two best results of each column are highlighted in bold and underline.}
\label{performance_conic40x}
\begin{tabular}{lcccccc}
\toprule
\textbf{Method} & \textbf{mPQ} & \textbf{mDQ} & \textbf{mSQ} &  \textbf{bPQ} & \textbf{Dice} & \textbf{Jacard}  \\
\midrule
HoverNet & 0.439 & 0.568 & 0.643 & 0.608  & 0.769 & 0.673\\
CiscNet & 0.503 & \textbf{0.646} & 0.694 & 0.628 & 0.773 & 0.678 \\
PointNu-Net & 0.495 & 0.644 & 0.665 & 0.612 & 0.734 & 0.651 \\
$\mathrm{CellViT_{SAM-L}}$ & \underline{0.517} & 0.624 & \underline{0.723} & \underline{0.656} & 0.782 & 0.692  \\
$\mathrm{CellViT_{UNI}}$ & 0.500 & 0.610 & 0.711 & 0.643 & \underline{0.783} & \underline{0.693}\\
$\mathrm{CellVTA}$ (ours) & \textbf{0.538} & \underline{0.645} & \textbf{0.737} & \textbf{0.675} & \textbf{0.797} & \textbf{0.714} \\


\bottomrule
\end{tabular}
\end{table}

\begin{table}[t]
\centering
\setlength{\tabcolsep}{2pt}
\caption{Performance (PQ) of difference methods on CoNIC across cell types.}
\label{performance_conic40x_detail}
\begin{tabular}{lcccccc}
\toprule
\textbf{Method} & \textbf{Neutro.} & \textbf{Epithelial} & \textbf{Lymph.} &  \textbf{Plasma} & \textbf{Eosin.} & \textbf{Connect.}  \\
\midrule
CiscNet & \textbf{0.269} & 0.584 & 0.578 & \underline{0.377} & \underline{0.316} & 0.549 \\
$\mathrm{CellViT_{SAM-L}}$ & 0.219 & \underline{0.600} & \underline{0.631} & 0.375 & 0.297 & \underline{0.568}  \\
$\mathrm{CellViT_{UNI}}$ & 0.184 & 0.594 & 0.602 & 0.350 & 0.308 & 0.542\\
$\mathrm{CellVTA}$ (ours)  & \underline{0.257} & \textbf{0.623} & \textbf{0.645} & \textbf{0.378} & \textbf{0.317} & \textbf{0.594} \\


\bottomrule
\end{tabular}
\end{table}

\textbf{Implementation Details.} The hyperparameters of CellVTA, CellViT and $\mathrm{CellViT_{UNI}}$ are based on the configuration in ~\cite{horst2024cellvit}. We use UNI~\cite{chen2024towards} as the backbone model. It can be easily replaced by any other VFM. During training of $\mathrm{CellViT_{UNI}}$ and CellVTA, we freeze the ViT encoder and only train the adapter and decoder.  We use AdamW~\cite{loshchilov2018decoupled} optimizer and incorporate exponential learning rate scheduling with a scheduling factor of 0.85. The initial learning rate is 3e-4 and the batch size is 4. We train our model for 50 epochs on CoNIC and 100 epochs on PanNuke. All experiments are conducted on a 32GB V100 GPU.

\textbf{Evaluation.} To quantitatively assess nuclei instance segmentation, we use dice coefficient (DICE), aggregated Jaccard index (AJI), binary panoptic quality (bPQ), and multi-class panoptic quality (mPQ) as metrics. Panoptic quality (PQ)~\cite{kirillov2019panoptic} consists of detection quality (DQ) and segmentation quality (SQ). For the CoNIC dataset, we apply an upsampling strategy during training and test, since we found that all models perform better at $40\times$ magnification. Each $256\times256$ px image is upsampled to $480\times480$ px by linear interpolation and split into 4 overlapping $256\times256$ px patches (32 px overlap). During inference, the predictions are merged and downsampled to the original $20\times$ magnification for evaluation.



\begin{table}[t]
\setlength{\tabcolsep}{6pt}
\centering
\caption{Performance comparison between CellVTA and baselines on PanNuke.}
\label{performance_pannuke}
\begin{tabular}{lcccccc}
\toprule
\textbf{Method} & \textbf{mPQ} & \textbf{mDQ} & \textbf{mSQ} &  \textbf{bPQ} & \textbf{Dice} & \textbf{Jacard}  \\
\midrule
HoverNet & 0.443 & 0.535 & 0.686 & 0.638  & 0.795 & 0.714\\
CiscNet & 0.469 & 0.575 & 0.673 & 0.649 & \textbf{0.830} & 0.719 \\
PointNu-Net & 0.480 & 0.576 & 0.687 & 0.660 & 0.774 & 0.690 \\
$\mathrm{CellViT_{SAM-L}}$ & \underline{0.492} & 0.591 & 0.704 & \underline{0.662} & 0.808 & \underline{0.734}  \\
$\mathrm{CellViT_{UNI}}$ & 0.491 & \underline{0.593} & \underline{0.705} & 0.654 & 0.806 & 0.730\\
$\mathrm{CellVTA}$ (ours) & \textbf{0.506} & \textbf{0.605} & \textbf{0.715} & \textbf{0.668} & \underline{0.811} & \textbf{0.740} \\


\bottomrule
\end{tabular}
\end{table}

\begin{table}[t]
\centering
\setlength{\tabcolsep}{4pt}
\caption{Performance (PQ) of difference methods on PanNuke across cell types.}
\label{performance_pannuke_cell}
\begin{tabular}{lccccc}
\toprule
\textbf{Method} & \textbf{Neoplastic} & \textbf{Inflamm.} & \textbf{Connect.} &  \textbf{Dead} & \textbf{Epithelial}   \\
\midrule
PointNu-Net & 0.565 & 0.405 & 0.397 & 0.136 & 0.566 \\
$\mathrm{CellViT_{SAM-L}}$ & \underline{0.580} & \underline{0.428} & 0.408 & 0.146 & \underline{0.578}  \\
$\mathrm{CellViT_{UNI}}$ & 0.572 & 0.427 & \underline{0.421} & \underline{0.179} & 0.572\\
$\mathrm{CellVTA}$ (ours)  & \textbf{0.585} & \textbf{0.434} & \textbf{0.433} & \textbf{0.185} & \textbf{0.592} \\


\bottomrule
\end{tabular}
\end{table}

\subsection{Results}
\noindent\textbf{Comparison with SOTA Methods.} We compare our method with the state-of-the-art methods, including HoverNet~\cite{graham2019hover}, CiscNet~\cite{bohland2022ciscnet}, PointNu-Net~\cite{yao2023pointnu}, and CellViT~\cite{horst2024cellvit}. The former three methods are representative CNN-based methods and CellViT is a SOTA ViT-based method. For CellViT, we use the original version with SAM-L~\cite{kirillov2023segment} as the encoder and a modified version with UNI~\cite{chen2024towards} as the encoder. As shown in Table~\ref{performance_conic40x}, CellVTA significantly outperforms the baseline models on CoNIC, improving the mPQ, bPQ, and Jaccard scores by 2.1\%, 1.9\%, and 2.1\% above the second-best method. CNN-based methods like CiscNet and PointNu-Net achieve higher mDQ scores than ViT-based methods. However, CellVTA significantly improves mDQ compared to CellViT and obtains comparable results with the SOTA CNN method, which indicates that the adapter improves the detection rate of ViT. We further compare the performance (PQ score) on each cell type for ViT-based methods and CiscNet which is the best CNN-based method. As shown in Table~\ref{performance_conic40x}, CellVTA achieves the best performance on 5 out of 6 cell types. For PanNuke, the performance is shown in Table~\ref{performance_pannuke} and Table~\ref{performance_pannuke_cell}. CellVTA outperforms all methods across all metrics and cell types, except for the Dice score, where CiscNet performs somewhat better. Overall, CellVTA consistently surpasses SOTA methods on both datasets, which shows the effectiveness of the adapter in leveraging the power of pathology foundation models.

\begin{figure}[t]
\includegraphics[width=0.98\textwidth]{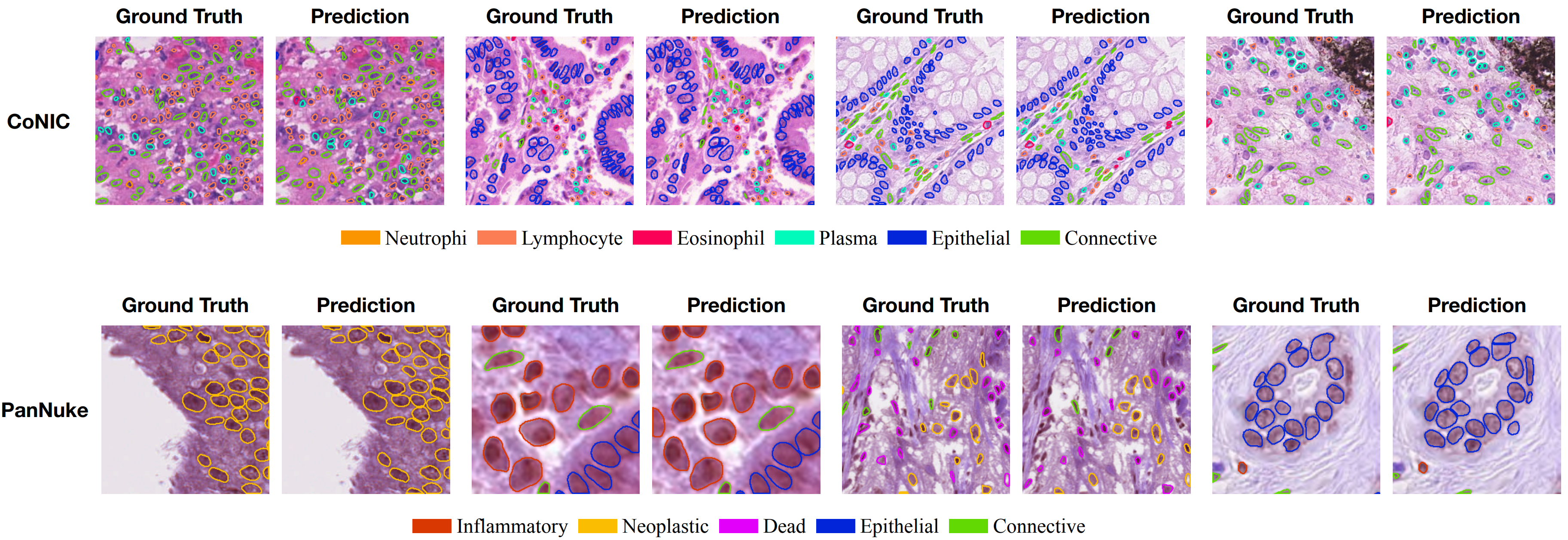}
\caption{Example of CoNIC and PanNuke patches with ground-truth annotations (left) and CellVTA predictions overlaid (right).} \label{qualitative_result}
\end{figure}

\textbf{Qualitative Results.} Fig.~\ref{qualitative_result} displays some segmentation results of CellVTA on two datasets. The cells in CoNIC are notably smaller than those in PanNuke. Additionally, cell sizes vary widely across cell types and tissue origins. Despite these challenges, CellVTA consistently produces high-quality segmentation and classification results, consistent with the ground truth, even for extremely small cells. Furthermore, despite the high heterogeneity in cellular composition in some images, our model is still able to accurately classify the majority of cells.

\begin{table}[t]
\setlength{\tabcolsep}{6pt}
\centering
\caption{Ablation studies on fine-tuning strategies. Full and Frozen mean full fine-tuning and freezing the encoder during training, respectively.}
\label{ablation_studies}
\begin{tabular}{lcccccc}
\toprule
\textbf{Method} & \textbf{mPQ} & \textbf{mDQ} & \textbf{mSQ} &  \textbf{bPQ} & \textbf{Dice} & \textbf{Jacard}  \\
\midrule
$\mathrm{CellViT_{SAM-L}}$-Full & 0.517 & 0.624 & 0.723 & 0.656  & 0.782 & 0.692\\
$\mathrm{CellViT_{SAM-L}}$-Frozen & 0.504 & 0.614 & 0.728 & 0.642 & 0.782 & 0.693 \\
$\mathrm{CellVTA_{SAM-L}}$ (ours) & \underline{0.526} & \underline{0.632} & \underline{0.736} & \underline{0.668} & \underline{0.793} & \underline{0.708} \\
$\mathrm{CellViT_{UNI}}$-Full & 0.516 & 0.623 & 0.726 & 0.656 & 0.787 & 0.698  \\
$\mathrm{CellViT_{UNI}}$-Frozen  & 0.500 & 0.610 & 0.711 & 0.643 & 0.783 & 0.693\\
$\mathrm{CellVTA_{UNI}}$ (ours) & \textbf{0.538} & \textbf{0.645} & \textbf{0.737} & \textbf{0.675} & \textbf{0.797} & \textbf{0.714} \\

\bottomrule
\end{tabular}
\end{table}

\textbf{Ablation Studies.} Table~\ref{ablation_studies} presents the results of the ablation studies. It shows that our method significantly outperforms decoder-only fine-tuning and even surpasses full fine-tuning, regardless of whether SAM or UNI is used as the backbone. This result highlights the effectiveness of our adapter module. Furthermore, the results suggest that pathology foundation models exhibit greater potential for cell segmentation compared to general vision foundation models.

\section{Conclusion}
Cell instance segmentation is a critical task in pathology image analysis. In this paper, we proposed a novel approach named CellVTA, which adds a CNN-based adapter to inject high-resolution spatial information into ViTs, alleviating its loss of detailed information. Extensive experiments have shown that our method effectively improves the performance of pathology foundation models in cell instance segmentation and outperforms the SOTA methods. Our research suggests that foundation models have great potential to explore in cell-level analysis.

\bibliographystyle{splncs04}
\bibliography{reference}

\begin{thebibliography}{10}
\providecommand{\url}[1]{\texttt{#1}}
\providecommand{\urlprefix}{URL }
\providecommand{\doi}[1]{https://doi.org/#1}

\bibitem{abraham2019novel}
Abraham, N., Khan, N.M.: A novel focal tversky loss function with improved attention u-net for lesion segmentation. In: 2019 IEEE 16th international symposium on biomedical imaging (ISBI 2019). pp. 683--687. IEEE (2019)

\bibitem{achiam2023gpt}
Achiam, J., Adler, S., Agarwal, S., Ahmad, L., Akkaya, I., Aleman, F.L., Almeida, D., Altenschmidt, J., Altman, S., Anadkat, S., et~al.: Gpt-4 technical report. arXiv preprint arXiv:2303.08774  (2023)

\bibitem{binder2021morphological}
Binder, A., Bockmayr, M., H{\"a}gele, M., Wienert, S., Heim, D., Hellweg, K., Ishii, M., Stenzinger, A., Hocke, A., Denkert, C., et~al.: Morphological and molecular breast cancer profiling through explainable machine learning. Nature Machine Intelligence  \textbf{3}(4),  355--366 (2021)

\bibitem{bohland2022ciscnet}
B{\"o}hland, M., Neumann, O., Schilling, M.P., Reischl, M., Mikut, R., L{\"o}ffler, K., Scherr, T.: Ciscnet-a single-branch cell nucleus instance segmentation and classification network. In: 2022 IEEE International Symposium on Biomedical Imaging Challenges (ISBIC). pp.~1--5. IEEE (2022)

\bibitem{chen2024towards}
Chen, R.J., Ding, T., Lu, M.Y., Williamson, D.F., Jaume, G., Song, A.H., Chen, B., Zhang, A., Shao, D., Shaban, M., et~al.: Towards a general-purpose foundation model for computational pathology. Nature Medicine  \textbf{30}(3),  850--862 (2024)

\bibitem{chen2023vision}
Chen, Z., Duan, Y., Wang, W., He, J., Lu, T., Dai, J., Qiao, Y.: Vision transformer adapter for dense predictions. In: The Eleventh International Conference on Learning Representations (2023)

\bibitem{dosovitskiy2021an}
Dosovitskiy, A., Beyer, L., Kolesnikov, A., Weissenborn, D., Zhai, X., Unterthiner, T., Dehghani, M., Minderer, M., Heigold, G., Gelly, S., Uszkoreit, J., Houlsby, N.: An image is worth 16x16 words: Transformers for image recognition at scale. In: International Conference on Learning Representations (2021)

\bibitem{gamper2020pannuke}
Gamper, J., Koohbanani, N.A., Benes, K., Graham, S., Jahanifar, M., Khurram, S.A., Azam, A., Hewitt, K., Rajpoot, N.: Pannuke dataset extension, insights and baselines. arXiv preprint arXiv:2003.10778  (2020)

\bibitem{graham2024conic}
Graham, S., Vu, Q.D., Jahanifar, M., Weigert, M., Schmidt, U., Zhang, W., Zhang, J., Yang, S., Xiang, J., Wang, X., et~al.: Conic challenge: Pushing the frontiers of nuclear detection, segmentation, classification and counting. Medical image analysis  \textbf{92},  103047 (2024)

\bibitem{graham2019hover}
Graham, S., Vu, Q.D., Raza, S.E.A., Azam, A., Tsang, Y.W., Kwak, J.T., Rajpoot, N.: Hover-net: Simultaneous segmentation and classification of nuclei in multi-tissue histology images. Medical image analysis  \textbf{58},  101563 (2019)

\bibitem{horst2024cellvit}
H{\"o}rst, F., Rempe, M., Heine, L., Seibold, C., Keyl, J., Baldini, G., Ugurel, S., Siveke, J., Gr{\"u}nwald, B., Egger, J., et~al.: Cellvit: Vision transformers for precise cell segmentation and classification. Medical Image Analysis  \textbf{94},  103143 (2024)

\bibitem{kirillov2019panoptic}
Kirillov, A., He, K., Girshick, R., Rother, C., Doll{\'a}r, P.: Panoptic segmentation. In: Proceedings of the IEEE/CVF conference on computer vision and pattern recognition. pp. 9404--9413 (2019)

\bibitem{kirillov2023segment}
Kirillov, A., Mintun, E., Ravi, N., Mao, H., Rolland, C., Gustafson, L., Xiao, T., Whitehead, S., Berg, A.C., Lo, W.Y., et~al.: Segment anything. In: Proceedings of the IEEE/CVF International Conference on Computer Vision. pp. 4015--4026 (2023)

\bibitem{liu2021swin}
Liu, Z., Lin, Y., Cao, Y., Hu, H., Wei, Y., Zhang, Z., Lin, S., Guo, B.: Swin transformer: Hierarchical vision transformer using shifted windows. In: Proceedings of the IEEE/CVF international conference on computer vision. pp. 10012--10022 (2021)

\bibitem{loshchilov2018decoupled}
Loshchilov, I., Hutter, F.: Decoupled weight decay regularization. In: International Conference on Learning Representations (2019)

\bibitem{raghu2021vision}
Raghu, M., Unterthiner, T., Kornblith, S., Zhang, C., Dosovitskiy, A.: Do vision transformers see like convolutional neural networks? Advances in neural information processing systems  \textbf{34},  12116--12128 (2021)

\bibitem{raza2019micro}
Raza, S.E.A., Cheung, L., Shaban, M., Graham, S., Epstein, D., Pelengaris, S., Khan, M., Rajpoot, N.M.: Micro-net: A unified model for segmentation of various objects in microscopy images. Medical image analysis  \textbf{52},  160--173 (2019)

\bibitem{stringer2024transformers}
Stringer, C., Pachitariu, M.: Transformers do not outperform cellpose. bioRxiv pp. 2024--04 (2024)

\bibitem{vadori2025mind}
Vadori, V., Peruffo, A., Gra{\~A}{\NG}c, J.M., Finos, L., Grisan, E.: Mind the gap: Evaluating patch embeddings from general-purpose and histopathology foundation models for cell segmentation and classification. arXiv preprint arXiv:2502.02471  (2025)

\bibitem{vaswani2017attention}
Vaswani, A.: Attention is all you need. Advances in Neural Information Processing Systems  (2017)

\bibitem{wang2021pyramid}
Wang, W., Xie, E., Li, X., Fan, D.P., Song, K., Liang, D., Lu, T., Luo, P., Shao, L.: Pyramid vision transformer: A versatile backbone for dense prediction without convolutions. In: Proceedings of the IEEE/CVF international conference on computer vision. pp. 568--578 (2021)

\bibitem{wang2022cell}
Wang, Y., Wang, Y.G., Hu, C., Li, M., Fan, Y., Otter, N., Sam, I., Gou, H., Hu, Y., Kwok, T., et~al.: Cell graph neural networks enable the precise prediction of patient survival in gastric cancer. npj Precision Oncology  \textbf{6}(1), ~45 (2022)

\bibitem{yao2023pointnu}
Yao, K., Huang, K., Sun, J., Hussain, A.: Pointnu-net: Keypoint-assisted convolutional neural network for simultaneous multi-tissue histology nuclei segmentation and classification. IEEE Transactions on Emerging Topics in Computational Intelligence  \textbf{8}(1),  802--813 (2023)

\bibitem{zhu2021deformable}
Zhu, X., Su, W., Lu, L., Li, B., Wang, X., Dai, J.: Deformable-detr: Deformable transformers for end-to-end object detection. In: International Conference on Learning Representations (2021)

\end{thebibliography}
%




\end{document}